\documentclass[conference,letterpaper]{IEEEtran}
\IEEEoverridecommandlockouts
\usepackage{cite}
\usepackage{graphicx}
\usepackage{booktabs}
\usepackage{amsmath}
\usepackage{amssymb}
\usepackage{multirow}
\usepackage{textcomp}
\usepackage{xcolor}
\usepackage{pifont}
\usepackage{url}
\usepackage{orcidlink}
\def\BibTeX{{\rm B\kern-.05em{\sc i\kern-.025em b}\kern-.08em
    T\kern-.1667em\lower.7ex\hbox{E}\kern-.125emX}}

\newcommand{\yes}{\ding{51}}
\newcommand{\no}{}

\begin{document}
\bstctlcite{IEEEtranBSTCTL:nodash}

\title{Can Conversational Temporal Dynamics Improve Depression Detection in Dyads? A Preliminary Investigation in Multi-Modality Perspectives}

\author{\IEEEauthorblockN{Anonymous Authors}
\IEEEauthorblockA{\textit{Affiliations removed for} \\
\textit{double-blind review}\\
City, Country \\
email address or ORCID}
}

\author{
\IEEEauthorblockN{
Hanie Kang\scalebox{0.75}{\orcidlink{0009-0005-8023-9672}}\textsuperscript{1},
Huang-Cheng Chou\scalebox{0.75}{\orcidlink{0000-0003-2125-5689}}\textsuperscript{2},
Sudarsana Reddy Kadiri\scalebox{0.75}{\orcidlink{0000-0001-5806-3053}}\textsuperscript{2},
Shrikanth Narayanan\scalebox{0.75}{\orcidlink{0000-0002-1052-6204}}\textsuperscript{2}
}
\IEEEauthorblockA{
\textsuperscript{1}\textit{Thomas Lord Department of Computer Science, University of Southern California, USA}\\
\textsuperscript{2}\textit{Signal Analysis and Interpretation Laboratory (SAIL), University of Southern California, USA}
}
}

\maketitle

\begin{abstract}
Automatic depression detection from clinical interviews typically models the semantic content and acoustic characteristics of participant speech. 
However, the interactional \emph{timing} between the clinician and participant remains comparatively under-modeled. 
We investigate \emph{conversational temporal dynamics}, specifically dyadic turn-pair timing, as a primary modality fused with self-supervised encoders. 
Evaluated on the DAIC-WOZ dataset, we compare a compact 24-dimensional timing module against frozen WavLM-large and RoBERTa-large baseline detectors. 
This temporal module achieves the highest single-modality performance on the development set. 
Furthermore, a convex-weighted late fusion strategy improves overall performance to 0.804 and 0.669 macro-F1 on the development and test sets, respectively. 
The learned fusion effectively assigns zero weight to acoustics, demonstrating that conversational timing serves as a lightweight, interpretable complement for dyadic depression screening.
\end{abstract}


\section{Introduction}
Depression is among the most prevalent mental-health disorders worldwide~\cite{world2017depression}, and speech-based screening is widely studied as a low-cost complement to clinical assessment~\cite{valstar2013avec,valstar2016avec}. 
Most systems model either \emph{acoustics} (prosodic and spectral voice characteristics increasingly encoded by self-supervised models such as WavLM~\cite{chen2022wavlm}) or \emph{semantics} (the lexical content of what a participant says, often with contextual text encoders such as RoBERTa~\cite{liu2019roberta}).

A clinical interview, however, is fundamentally a \emph{dyadic} interaction: an interviewer (in DAIC-WOZ, the virtual interviewer ``Ellie'') asks, and the participant responds, in alternating turns. 
A growing body of work shows that the \emph{timing} of this exchange carries depression-relevant information. 
At the level of the individual speaker, depressed participants tend to respond more slowly, pause longer and more often within a turn, and speak at a reduced rate~\cite{yamamoto2020timing,mundt2012vocal,cummins2015review}. 
At the level of the exchange, turn-taking descriptors (floor control, turn-switch, and turn-hold offsets) carry mood-episode information beyond the acoustic signal itself~\cite{aldeneh2019mood}. 
Computational studies likewise find discriminative timing in session-level utterance/pause/response intervals~\cite{fushimi2026beyond} and durations between fine-grained acoustic landmarks~\cite{huang2020landmark,zhang2024llms}. 
We refer to these interaction-level timing signals as \emph{conversational temporal dynamics} (CTD). 
Because CTD describes the \emph{structure} of the exchange rather than its lexical content or short-frame spectral detail, it is a plausible candidate for complementary fusion.

Three recent lines of evidence converge on the same underlying claim that timing is discriminative, while each leaves the specific gap this paper targets. 
In \emph{deception} detection, which has the same dyadic Ask/Res structure as a clinical interview, Chou et al.~\cite{chou2019joint,chou2021automatic} show that compact dyadic turn-pair timing descriptors improve detection over acoustics alone, but they do not study depression. 
On DAIC-WOZ, Fushimi et al.~\cite{fushimi2026beyond} show that long-term utterance/pause/response intervals help, but treat timing as a \emph{single-speaker} acoustic feature. 
Zhang et al.~\cite{zhang2025speechtrag,zhang2024llms} likewise show that acoustic-landmark timing carries a depression signal that text-only models miss, but use it at the sub-phonetic level as a retrieval aid rather than as a modality in its own right. 
Across all three, timing helps; yet the \emph{dyadic, transcript-level turn structure} of the interview remains underexplored as a first-class modality fused with strong self-supervised encoders.

We therefore treat CTD as a standalone modality alongside frozen WavLM-large acoustics and frozen RoBERTa-large semantics, evaluating all three on DAIC-WOZ~\cite{gratch2014daic} under the same strict, leakage-safe, subject-independent protocol. 
The resulting study defines a compact, interpretable 24-dimensional CTD module from Ask/Res turn pairs, compares A, T, CTD, and their pairwise/three-way late fusions, and asks how much weight conversational timing earns when placed on equal footing with 1024-dimensional neural encoders. 
On this small benchmark, CTD is the strongest single modality on the development subset, and convex-weighted fusion improves over the best single modality on both dev and test; the largest gain comes from adding CTD to semantics, while the learned three-way fusion assigns zero weight to acoustics. 
We treat these findings as preliminary and discuss the statistical caveats of the single small benchmark in Sec.~\ref{sec:discussion}. 
Our code and seed/hyperparameter/deployed-model details are available \footnote{\url{https://github.com/dndbsl/depression-detection-ctd}}.

\section{Related Work}
\label{sec:related}

\subsection{Acoustic and semantic depression detection on DAIC-WOZ}
The AVEC challenge series~\cite{valstar2013avec,valstar2016avec} established DAIC-WOZ-style audio/text depression assessment as a benchmark, and comprehensive reviews document the breadth of speech-based approaches to depression and related conditions~\cite{cummins2015review}. 
Early systems relied on hand-crafted prosodic and lexical features and their fusion~\cite{alhanai2018detecting}; more recent work adopts self-supervised speech encoders such as wav2vec 2.0~\cite{baevski2020wav2vec} and WavLM~\cite{chen2022wavlm}, and contextual text encoders such as RoBERTa~\cite{liu2019roberta}, typically probing frozen representations with a lightweight head in the SUPERB style~\cite{yang2021superb}. 
Because AVEC withheld the test labels, systems on DAIC-WOZ are conventionally compared on the development subset~\cite{gong2017topic,shen2022automatic,wu2023ssl,li2025hierarchical}, a convention we follow (Sec.~\ref{sec:protocol}); indeed, recent SSL systems report high dev macro-F1 under this convention, including WavLM/SSL ensemble scores of 0.800--0.829 and a WavLM-large hierarchical model at 0.81~\cite{wu2023ssl,li2025hierarchical}. 
Held-out test performance, when reported, is often more modest: recent speech-foundation-model studies on DAIC/E-DAIC report test or test-side average F1 scores in roughly the mid-0.50s to high-0.60s despite using WavLM, HuBERT, wav2vec 2.0, AudioMAE, or domain-adversarial training~\cite{dumpala2025ttt,kim2025dat}. 
We use frozen WavLM-large and RoBERTa-large as representative unimodal backbones; they are baselines here rather than contributions, since our question concerns what CTD adds on top of such encoders rather than how to maximize acoustic or semantic accuracy.

\subsection{Speech timing and dyadic interaction cues}
A long line of clinical and computational work links depression to \emph{when} and \emph{how much} a person speaks rather than only to voice quality. 
Psychomotor retardation, a core symptom of depression, manifests as slowed responses, longer and more frequent pauses, and reduced speech rate~\cite{yamamoto2020timing,mundt2012vocal}. 
Building on these observations, Fushimi et al.~\cite{fushimi2026beyond} propose session-level acoustic feature sets that explicitly summarize utterance, pause, and response intervals across an entire DAIC-WOZ session, and report that these long-term temporal descriptors, particularly pause-interval statistics, improve subject-independent classification when combined with conventional short-frame features, while contributing little on their own. 
A parallel line captures timing at a much finer, sub-phonetic granularity. For example, Huang et al.~\cite{huang2020landmark} model counts and durations of acoustic-\emph{landmark} n-grams, and Zhang et al.~\cite{zhang2024llms,zhang2025speechtrag} show that the durations between landmark pairs differ significantly between depressed and healthy speakers and can inject depression-relevant timing into a text-based large language model via retrieval. 
What these approaches share is a treatment of timing as a property of the \emph{participant's own signal}. The complementary view, that the \emph{coordination between the two interlocutors} is itself informative, is comparatively underexplored for depression. 
Specifically, Aldeneh et al.~\cite{aldeneh2019mood} show that dyadic turn-taking features (floor control, turn-switch and turn-hold offsets) improve mood-episode detection in clinical interviews, motivating an explicit model of the interviewer-participant exchange.

\subsection{Conversational temporal dynamics as a feature set}
Our CTD module operationalizes the dyadic view with a fixed, previously published descriptor set. 
The formulation originates in the deception-detection work of Chou et al.~\cite{chou2019joint,chou2021automatic}, who segment a dyadic interaction into consecutive Ask/Res (question/answer) turn pairs and, for each pair, compute turn durations, their differences, sums and ratios, voiced-versus-silence ratios for each side, the response latency (hesitation), and backchannel/silence counts. 
Their results suggest that turn-level timing is not redundant with voice acoustics and that timing from both sides of an exchange can be discriminative. Because a clinical interview shares this Ask/Res structure (Ellie asks, the participant responds), it is natural to test whether the same dyadic timing descriptors transfer to depression. We reuse the descriptor set essentially unchanged and as a fixed, task-agnostic feature bank, so that any predictive value on DAIC-WOZ reflects the cues themselves rather than dataset-specific feature engineering.

\subsection{Multimodal fusion for depression} 
Text--audio (and audio--visual) fusion for depression is well studied at the architecture level, spanning early/feature concatenation, late/decision fusion, and attention-based fusion~\cite{alhanai2018detecting}. 
A recurring difficulty is dimensionality imbalance: a low-dimensional, interpretable descriptor set concatenated with high-dimensional neural embeddings tends to be swamped by the latter. Prior timing work sidesteps this by either fusing timing with only one other stream~\cite{chou2019joint,fushimi2026beyond} or injecting it through a separate mechanism such as retrieval~\cite{zhang2025speechtrag}. We instead keep every per-modality detector and the evaluation protocol fixed and vary only the modality \emph{set} and the score-level fusion rule, fusing at the \emph{score} level so that a 24-dimensional timing module and 1024-dimensional neural encoders each contribute a single session probability on equal footing. This isolates the effect of \emph{adding CTD} from any change in encoder or protocol, and lets a dev-tuned convex combination reveal how much weight each modality actually earns.

\begin{figure*}[!t]
\centering
\includegraphics[width=\textwidth]{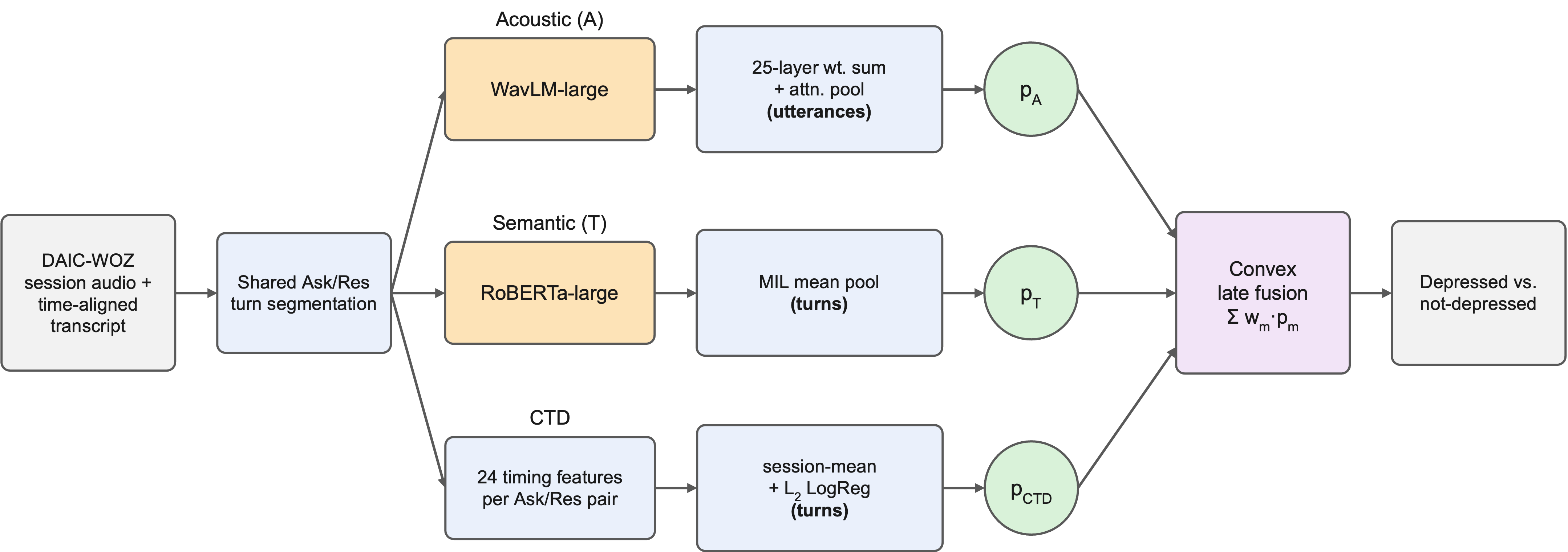}
\caption{Tri-modal pipeline. All three detectors share one Ask/Res turn segmentation of the same interview, but differ in \emph{instance granularity}: the frozen WavLM-large acoustic branch pools over participant \emph{utterances}, while the RoBERTa-large semantic branch and the 24-d CTD branch pool over participant \emph{turns}. Each branch emits one session-level depression probability ($p_A, p_T, p_{\mathrm{CTD}}$); score-level convex fusion combines them with dev-tuned weights and a dev-tuned threshold. The learned three-way weights $(0.0,\,0.3,\,0.7)$ drive the acoustic stream to zero, so the deployed system reduces to RoBERTa+CTD.}
\label{fig:pipeline}
\vspace{-4mm}
\end{figure*}

\section{Resource}
\label{sec:data}
We use DAIC-WOZ~\cite{gratch2014daic}, which consists of semi-structured clinical interviews conducted by an animated virtual interviewer (``Ellie'') along with audio, time-aligned transcripts, and PHQ-8~\cite{kroenke2009phq8} depression labels binarized at PHQ-8 $\geq 10$ (depressed). 
DAIC-WOZ has been treated as a benchmark for depression assessment for nearly a decade. However, it is known to be small, noisy, and vulnerable to reproducibility and leakage pitfalls~\cite{danylenko2026pitfalls}. 
We therefore use it as a controlled benchmark rather than as evidence of clinical deployment readiness. These labels are self-report screening scores rather than clinical diagnoses, so we treat the task as depression \emph{screening} throughout. 
We use the official subject-independent train/dev/test split. Ten sessions are excluded ($\{318, 321, 341, 362, 373, 409, 444, 451, 458, 480\}$) because they are the sessions with documented data-integrity problems cataloged by Patapati~\cite{patapati2024integrating}, such as transcript or audio desynchronization, long non-interview interruptions, missing interviewer utterances, and a PHQ-8 or binary-label mismatch. 
We manually inspected these sessions against the catalog and, where that work manually \emph{repairs} them, take the more conservative route and \emph{exclude} them so that no manually reconstructed data enters training or evaluation. 
The single label-dependent exclusion (409) follows this published integrity catalog rather than our own inspection of the labels, ensuring that no held-out information is used. 
A shared cleaning step then removes empty or short ($<100$\,ms) participant utterances. This leaves 180 sessions. 
After applying the official split and aligning across all three modalities, we obtain 102 train, 33 dev, and 45 test sessions, with 29, 12, and 14 depressed participants respectively, representing about 30\% prevalence. 
All modalities are derived from the same transcript-based Ask/Res turn segmentation so that the three detectors see the same conversational units.

\section{Baselines}
\label{sec:modalities}
Each modality is an independently trained, deployable detector that outputs a per-session depression probability. 
The acoustic and semantic branches are intentionally \emph{standard baselines} consisting of frozen self-supervised encoders (WavLM-large, RoBERTa-large) with lightweight probing heads, following established practice~\cite{yang2021superb}. 
These are configured to a reasonable operating point rather than exhaustively optimized. 
They are not themselves a contribution, as our claims concern whether CTD adds value \emph{on top of} such strong, representative unimodal detectors. 
Therefore, for these two branches, we report the design we used without treating every hyperparameter as a research question. We describe each in turn, and Fig.~\ref{fig:pipeline} gives the overall pipeline.

\subsection{Acoustic (A): Frozen WavLM-Large}
We extract frozen \texttt{microsoft/wavlm-large} hidden states for every participant utterance and cache them as per-layer, time-mean-pooled vectors of shape $[\text{utterances}, 25, 1024]$ (the embedding output plus 24 transformer layers). 
A deliberately tiny SUPERB-style probe~\cite{yang2021superb} is trained on top: (i) a learnable softmax weighting over the 25 layers collapses the layer axis to one 1024-d vector per utterance; (ii) additive attention pooling with a single learnable query ($\text{score}_i = v^\top \tanh(W u_i)$, hidden size 256, masked over padding) aggregates utterances into a 1024-d session embedding $e_a$; (iii) dropout (0.3) and a single $\text{Linear}(1024 \to 1)$ produce the logit. The probe has roughly $0.5$M trainable parameters; WavLM is frozen and used only during one-time feature extraction. Training uses AdamW (lr $5\times10^{-5}$, weight decay $10^{-3}$), label smoothing 0.1, class-balanced positive weighting, batch size 32, up to 200 epochs without early stopping, and selects the lowest-dev-loss checkpoint. The session embedding $e_a$ is also exported for late fusion.

\noindent\textbf{Instance granularity: utterance versus turn.} 
The acoustic detector pools over participant \emph{utterances} (individual transcript rows), whereas the semantic and CTD modules pool over participant \emph{turns} (consecutive participant runs that answer one Ellie question, which is the same Ask/Res segmentation used for CTD). 
We tested an Ask/Res \emph{turn}-level acoustic variant using an identical probe, but with WavLM features time-mean-pooled over each whole response turn rather than per utterance. 
This approach was clearly worse, achieving a dev macro-F1 of 0.572 and a test macro-F1 of 0.333 at a threshold of 0.5, compared to 0.673 and 0.545 for the utterance-level model. 
This variant collapsed toward the majority class on the test set, misclassifying 24 out of 31 negatives. 
We attribute this outcome to the timescale of the acoustic cue. 
Depression-relevant prosody and voice quality reside at the sub-utterance level, so mean-pooling WavLM states over an entire multi-utterance turn averages away precisely the short-term spectral variation the probe relies on. 
This further yields fewer, longer, and more heterogeneous bags for the attention pooler to weigh. 
While turn boundaries carry the timing and lexical signal that the CTD and semantic modules exploit, they are not where the acoustic signal concentrates. 
We therefore keep the acoustic detector at utterance granularity and let the CTD model the turn structure explicitly.

\subsection{Semantic (T): Frozen RoBERTa-Large}
Each participant turn (Ellie rows define boundaries; only participant language is retained, up to 512 tokens) is encoded with a frozen \texttt{roberta-large} and mean-pooled over real tokens into a 1024-d instance embedding. 
Per-feature standardization (statistics fit on the training data only) is applied and stored in the checkpoint. 
Instances are aggregated per participant by multiple-instance-learning (MIL) mean pooling, followed by dropout (0.1) and a $\text{Linear}(1024 \to 1)$ head. 
We use non-parametric mean pooling, a standard MIL aggregator that adds no parameters and is a robust default at this dataset scale ($\approx 57$ turns per participant, $\sim$100 training sessions). Only the classifier head is trained (AdamW, lr $10^{-3}$, weight decay 0.01, BCE with train-split positive weighting, 200 epochs, effective batch size 32), selecting the checkpoint with the highest dev macro-F1. Participants average $\approx 57$ turns of $\approx 25$ words.

\subsection{Conversational Temporal Dynamics (CTD)}
\label{sec:ctd}

\begin{figure}[!t]
\centering
\includegraphics[width=\columnwidth]{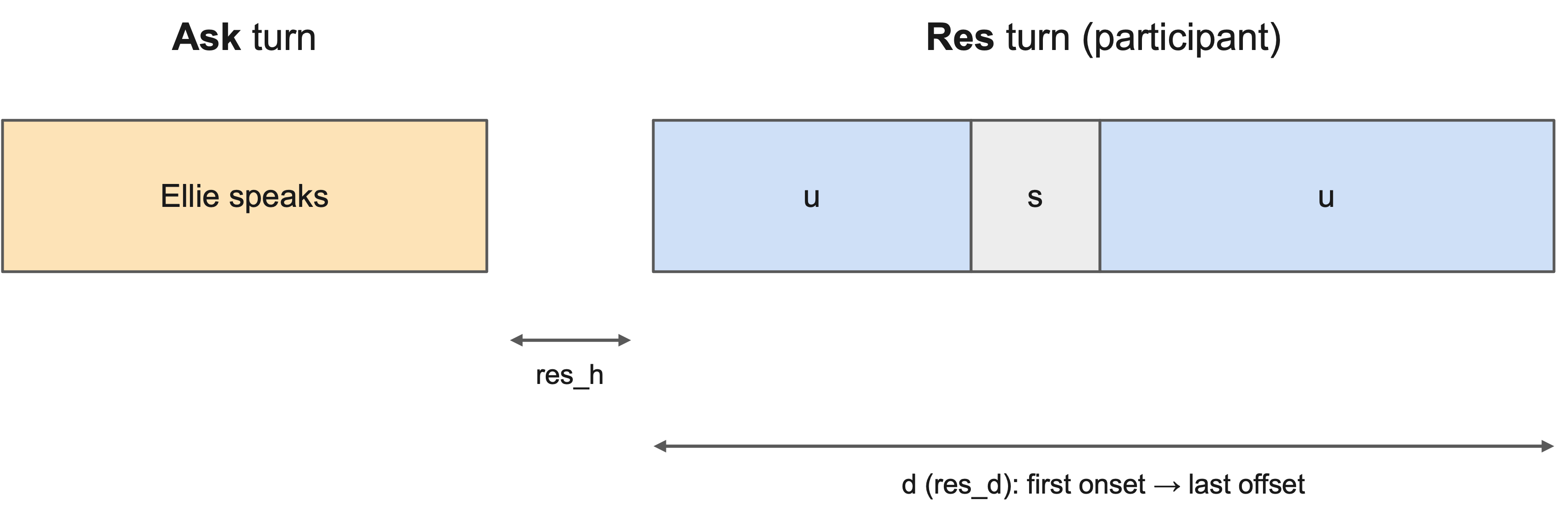}
\caption{One Ask/Res turn pair. For each turn we measure duration $d$, voiced time $u$ (blue), and within-turn silence $s{=}d{-}u$ (grey); \texttt{res\_h} is the response latency (Ask-end $\to$ Res-start gap). The 24 CTD features (Table~\ref{tab:ctd}) are pairwise ratios/differences among $\{d,u,s\}$ for the Ask and Res sides, plus \texttt{res\_h} and backchannel/silence counts, averaged over a session.}
\label{fig:ctd}
\vspace{-4mm}
\end{figure}

The CTD module describes the \emph{timing structure} of the dyadic exchange (Fig.~\ref{fig:ctd}). 
Rather than engineer a new feature set on DAIC-WOZ, we \emph{adopt a fixed, previously published} dyadic timing descriptor set originally designed for a different task and corpus (Sec.~\ref{sec:related}). 
This set is \emph{a priori} and task-agnostic, as it is neither tuned nor added to or removed from based on DAIC-WOZ performance, thereby keeping the feature design leakage-safe by construction. 
The original formulation specifies its descriptors on the Ask side and mirrors them on the Res side. Realizing this specification symmetrically for both interlocutors yields the 24 features in Table~\ref{tab:ctd}. 
We segment each transcript into alternating interviewer (Ask, ``Ellie'') and participant (Res) speaker runs, forming consecutive Ask/Res \emph{turn pairs}. 
For each turn pair, we compute the 24 timing descriptors, grouped into turn durations, their differences, sums, and ratios, twelve voiced-versus-silence ratio features for the Ask and Res turns, the response latency (hesitation), and backchannel or within-turn-silence counts. 
Concretely, for a turn, we define its \emph{duration} $d$ (first-onset to last-offset), \emph{voiced} time $u$ (sum of utterance lengths), and \emph{silence} time $s = d - u$. 
The ratio features are the pairwise quotients among $\{d, u, s\}$ for Ask and Res, and the hesitation $\text{res\_h}$ is the gap between the end of the Ask turn and the start of the Res turn. 
Undefined ratios (e.g., zero-silence turns) result in NaN and are imputed with train-split medians, ensuring a leakage-safe process. 
We use interviewer turns only as timing anchors and never feed their lexical content to the semantic detector, in light of evidence that DAIC-WOZ interviewer prompts can create textual shortcuts~\cite{burdisso2024prompts}. 
Whereas the original work feeds the per-turn-pair sequence to a recurrent network, we deliberately keep the module minimal for this small cohort.

For the deployed CTD detector, we take the per-session, NaN-aware mean of each of the 24 features (a 24-d session vector), standardize (train-fit), and classify with an $L_2$-regularized logistic regression ($C{=}0.3$, class-balanced).\label{sec:ctd-deploy} 
Following the challenge convention, dev probabilities come from a train-only fit and test probabilities from a train+dev refit. This session-mean summary is our default \emph{before} looking at results; the only design freedom is the classifier and its regularization, not the feature set. 
As a post-hoc robustness check we also implemented a richer variant that summarizes each feature's per-turn \emph{sequence} with the 10-operator eGeMAPS functional bank~\cite{eyben2016egemaps} (mean, coefficient of variation, 20/50/80th percentiles, 20--80 range, and rising/falling slope mean/std), giving $24\times10=240$ descriptors and selecting among logistic regression, SVM, and random forest by dev balanced accuracy; this more expressive variant did \emph{not} beat the compact 24-d session-mean detector on dev, so we retain the simpler pre-committed model for fusion.

\begin{table}[!t]
\centering
\caption{The 24 conversational temporal dynamics (CTD) features, computed per Ask/Res turn pair and averaged over a session. $d$: turn duration; $u$: voiced time; $s{=}d{-}u$: within-turn silence.}
\label{tab:ctd}
\fontsize{7}{9}\selectfont
\setlength{\tabcolsep}{3pt}
\vspace{-2mm}
\begin{tabular}{ll}
\toprule
Group ($\#$) & Features \\
\midrule
Durations (2) & \texttt{ask\_d}, \texttt{res\_d} \\
Diff/sum (3) & \texttt{res\_minus\_ask}, \texttt{ask\_minus\_res}, \texttt{duration\_sum} \\
Duration ratios (2) & \texttt{res\_over\_ask}, \texttt{ask\_over\_res} \\
Ask voiced/silence (6) & \texttt{ask\_ud}, \texttt{ask\_du}, \texttt{ask\_sd}, \texttt{ask\_ds}, \texttt{ask\_su}, \texttt{ask\_us} \\
Res voiced/silence (6) & \texttt{res\_ud}, \texttt{res\_du}, \texttt{res\_sd}, \texttt{res\_ds}, \texttt{res\_su}, \texttt{res\_us} \\
Hesitation (1) & \texttt{res\_h} (Ask-end $\to$ Res-start gap) \\
Counts (4) & \texttt{ask\_bt}, \texttt{res\_bt} (overlap), \texttt{ask\_st}, \texttt{res\_st} (silence) \\
\bottomrule
\end{tabular}
\vspace{-4mm}
\end{table}

\section{Methodology}
\vspace{-1mm}
\label{sec:fusion}
Because the three detectors have very different input dimensionalities, we fuse at the \emph{score} (probability) level so each modality contributes a single session score. Let $p_m \in [0,1]$ be modality $m$'s session probability and $S$ be a modality subset. The parameter-free \texttt{mean\_prob} rule averages the available probabilities:
\[
p_S^{\mathrm{mean}}
= \frac{1}{|S|}\sum_{m\in S} p_m .
\]
We also implemented an equal-weight log-odds variant,
\[
p_S^{\mathrm{logit}}
= \sigma\!\left(
\frac{1}{|S|}\sum_{m\in S}\operatorname{logit}(p_m)
\right),
\]
but it behaved near-identically to \texttt{mean\_prob} in earlier runs and is omitted from the result table for space. Our dev-tuned rule is a convex combination,
\[
p_S^{\mathrm{w}}
= \sum_{m\in S} w_m p_m,
\quad
w_m \geq 0,\quad
\sum_{m\in S} w_m = 1,
\]
where weights are grid-searched on dev with step 0.1. For every fusion rule the decision threshold is tuned on dev to maximize dev macro-F1; for \texttt{wconvex}, both the weights and the threshold are chosen on dev. Single-modality detectors use their own deployed thresholds (WavLM 0.65, RoBERTa 0.524, CTD 0.50). The convex-weighted fusion of all three modalities is our proposed system.

\noindent\textbf{Threshold selection.} Every threshold in this paper is chosen on dev alone (grid $0.05$--$0.95$, step $0.05$, maximizing dev macro-F1) and then applied unchanged to test, so test never informs any decision boundary. For the single detectors this dev-tuned operating point is stored with the checkpoint: RoBERTa's is 0.524 and CTD's is 0.50. The acoustic threshold is \emph{not} 0.50: because the WavLM checkpoint is selected by lowest dev loss rather than by F1, its logits are not centered for balanced decisions, and its dev-optimal threshold is 0.65 (at the default 0.50 the same checkpoint scores only test 0.399). Class imbalance ($\approx$30\% positive) likewise pushes the F1-optimal threshold above 0.50. For fusion, the threshold is re-tuned on dev jointly with the convex weights, since averaging modality probabilities changes the score distribution.

\section{Experimental Setup and Evaluation}
\vspace{-1mm}
\label{sec:protocol}
We follow a strict, leakage-safe, subject-independent protocol. No participant appears in more than one split; all preprocessing statistics, model parameters, fusion weights, and thresholds are fit or selected using train and/or dev only. \textbf{Reporting.} Following prior work on DAIC-WOZ~\cite{gratch2014daic,valstar2013avec,alhanai2018detecting,wu2023ssl,li2025hierarchical} -- where the AVEC test labels were originally withheld and systems are therefore commonly compared on the development subset -- we report results \emph{primarily on dev}, with macro-F1 as the headline metric, and report held-out test for reference. This convention is imperfect but common: a recent DAIC-WOZ methodological review found that only a minority of studies report held-out test performance at all~\cite{danylenko2026pitfalls}. For the test report each detector is refit on train+dev (the standard ``deploy on all development data'' step) and evaluated \emph{once}; we never select on test. We compute participant-level 95\% confidence intervals from 2{,}000 bootstrap resamples~\cite{ferrer2024confidence} for both dev and test. Because the same participants are evaluated by each detector, we also compute paired bootstrap confidence intervals for macro-F1 \emph{differences} between the best fusion and its component single modalities, plus prediction-change counts. Because dev ($n{=}33$) is small and the convex weights are dev-tuned, dev numbers for \textbf{wconvex} are mildly optimistic; we therefore also report the parameter-free \textbf{mean\_prob} fusion as a no-tuning reference.

\section{Experimental Results and Analyses}
\label{sec:results}

\subsection{Single modalities} 
Among the unimodal detectors (Table~\ref{tab:results}, top), the \textbf{CTD module is the strongest single modality on dev} (dev macro-F1 0.746, test 0.631), ahead of frozen RoBERTa-large semantics (0.690 / 0.631) and frozen WavLM-large acoustics (0.673 / 0.545). That a 24-dimensional, fully interpretable timing module matches or beats 1024-dimensional self-supervised encoders is the central empirical motivation for treating CTD as a modality in its own right. On test, CTD and RoBERTa tie at 0.631 macro-F1, while WavLM trails at 0.545.

\subsection{Fusion improves, and CTD drives the gain} 
Our proposed convex-weighted late fusion improves over the best single modality on both splits: \textbf{0.804} dev macro-F1 ($+0.058$ over CTD's 0.746) and \textbf{0.669} test macro-F1 ($+0.038$ over 0.631) (Table~\ref{tab:results}, bottom; the bold row is our proposed three-way system). The improvement is attributable specifically to CTD. Adding CTD to the semantic detector is the single most effective step -- it lifts dev macro-F1 from 0.690 to 0.804 ($+0.114$) and test from 0.631 to 0.669 ($+0.038$). In contrast, \emph{no} fusion that includes the acoustic stream improves test macro-F1 over the best single modality. Strikingly, when the convex weights are learned over all three modalities, the acoustic weight is driven to \textbf{exactly zero} ($w = [\,\text{A}{=}0.0,\ \text{T}{=}0.3,\ \text{CTD}{=}0.7\,]$), so the three-way system recovers the RoBERTa+CTD detector: under this dev-tuned grid, the acoustic branch receives no weight.

\subsection{Fusion-rule comparison} 
Table~\ref{tab:strategies} compares parameter-free equal weighting (\texttt{mean\_prob}) with dev-tuned convex weighting (\texttt{wconvex}) across modality subsets. Equal weighting of all three modalities reaches dev macro-F1 0.710 but generalizes poorly to test (0.545), because it forces a 1/3 weight on the unhelpful acoustic stream; the dev-tuned convex weighting instead down-weights or removes acoustics and generalizes better (test 0.669). The parameter-free \texttt{mean\_prob[RoBERTa+CTD]} (dev 0.738 / test 0.650) is a strong no-tuning fallback that already beats every single modality on test, indicating the CTD-with-semantics gain is not merely an artifact of dev-tuning the weights.

\begin{table}[!t]
\centering
\caption{Depression detection on DAIC-WOZ. A: WavLM-large; T: RoBERTa-large; CTD: conversational temporal dynamics. \yes{} marks active modalities. Single rows use the deployed detectors at their own thresholds; multi-modality rows use convex-weighted late fusion (weights + threshold tuned on dev). Dev macro-F1 is the primary metric; test is reported for reference; both include 95\% bootstrap CIs. Our proposed system (three-way fusion) is in \textbf{bold}; its learned acoustic weight is 0, so it coincides with RoBERTa+CTD.}
\label{tab:results}
\begin{tabular}{ccc l l}
\toprule
A & T & CTD & Dev F1$_{\text{m}}$ [95\% CI] & Test F1$_{\text{m}}$ [95\% CI] \\
\midrule
\yes & \no  & \no  & 0.673 [0.498, 0.818] & 0.545 [0.392, 0.681] \\
\no  & \yes & \no  & 0.690 [0.517, 0.833] & 0.631 [0.474, 0.769] \\
\no  & \no  & \yes & 0.746 [0.581, 0.879] & 0.631 [0.472, 0.771] \\
\midrule
\yes & \yes & \no  & 0.700 [0.520, 0.847] & 0.568 [0.410, 0.709] \\
\yes & \no  & \yes & 0.775 [0.618, 0.906] & 0.612 [0.458, 0.750] \\
\no  & \yes & \yes & 0.804 [0.643, 0.935] & 0.669 [0.509, 0.806] \\
\midrule
\textbf{\yes} & \textbf{\yes} & \textbf{\yes} & \textbf{0.804 [0.643, 0.935]} & \textbf{0.669 [0.509, 0.806]} \\
\bottomrule
\end{tabular}
\vspace{-4mm}
\end{table}

\subsection{Other metrics, paired uncertainty, and seed variability} 

The dev-selected best system (\texttt{wconvex[RoBERTa+CTD]}) reaches dev BA 0.804, test BA 0.673, test depressed-class F1 0.552, and test ROC-AUC 0.682. Thus, the macro-F1 gain in Table~\ref{tab:results} is not accompanied by a collapse in clinically relevant sensitivity/specificity balance or depressed-class performance. As expected for $n{=}33$ dev / $n{=}45$ test, the bootstrap CIs are wide and overlap across arms (e.g.\ test macro-F1 CI $[0.509, 0.806]$ for our system vs.\ $[0.472, 0.771]$ for CTD alone), so the test-side ordering should be read as indicative. Paired bootstrapping over the same predictions gives positive but non-definitive RoBERTa+CTD deltas relative to both RoBERTa and CTD; all paired intervals include or touch zero. At the prediction level, RoBERTa+CTD changes 18/45 test decisions relative to RoBERTa (10 corrected, 8 worsened) and 2/45 relative to CTD (both corrected). Existing neural seed sweeps show RoBERTa-large frozen test macro-F1 $0.604{\pm}0.025$ over seeds 42/43/44 and WavLM-large test macro-F1 $0.506{\pm}0.034$ over six seeds, placing the observed fusion gain in the same small-sample variability context. At the same time, the 0.650--0.669 test macro-F1 of the no-tuning and convex RoBERTa+CTD fusions is competitive with recent test-side speech-foundation-model results on DAIC/E-DAIC: Dumpala et al. report DAIC test macro-F1 around 0.60--0.67 for SSL models, and Kim et al. report E-DAIC depression F1 averages around 0.55--0.68 before/after gender domain-adversarial training~\cite{dumpala2025ttt,kim2025dat}. Thus, the consistent dev improvement, its reproduction by the parameter-free fusion, and the learned zero acoustic weight are the more robust observations, while the uncertainty analyses caution against claiming a definitive test-side win.

\section{Discussion and Limitations}
\vspace{-1mm}
\label{sec:discussion}
Three points stand out. First, a 24-dimensional, fully interpretable timing module is competitive with 1024-dimensional self-supervised acoustic and semantic encoders, and on the development subset it exceeds both, reinforcing that \emph{how the conversation is timed} may be as informative as \emph{how it sounds} or \emph{what is said}. Second, the complementarity is specific: CTD composes with \emph{semantics} (response content plus response timing), whereas the frozen WavLM acoustic stream contributes little on this cohort and receives zero weight in the learned three-way fusion -- consistent with the acoustic detector's own weak standalone test performance (0.545). Third, score-level fusion that places CTD on equal footing with the neural encoders is the natural way to exploit it; equal weighting that forces in the acoustic stream overfits dev, while dev-tuned convex weighting transfers to test.

\begin{table}[!t]
\centering
\caption{Fusion-rule comparison (dev / test macro-F1). \texttt{wconvex} weights are grid-searched on dev (shown as A/T/CTD). Forcing in the acoustic stream (A+T, A+T+CTD) hurts test under both rules; dev-tuned convex weighting, which can down-weight or drop acoustics, generalizes best.}
\label{tab:strategies}
\begin{tabular}{l cc l}
\toprule
Subset & mean\_prob & wconvex & wconvex weights \\
\midrule
A+T       & 0.646 / 0.525 & 0.700 / 0.568 & 0.2 / 0.8 / -- \\
A+CTD     & 0.746 / 0.537 & 0.775 / 0.612 & 0.1 / -- / 0.9 \\
T+CTD     & 0.738 / 0.650 & 0.804 / 0.669 & -- / 0.3 / 0.7 \\
A+T+CTD   & 0.710 / 0.545 & 0.804 / 0.669 & 0.0 / 0.3 / 0.7 \\
\bottomrule
\end{tabular}
\vspace{-4mm}
\end{table}

\textbf{Limitations.} DAIC-WOZ is small and class-imbalanced, and recent reviews emphasize that the benchmark is vulnerable to underdocumented protocols, subject leakage, and shortcut learning~\cite{danylenko2026pitfalls}. We nevertheless use it because it remains a common benchmark for this task and lets us compare CTD with the development-set reporting convention used by prior work. Consistent with prior work our primary comparison is on the development subset, which is itself small ($n{=}33$); held-out test numbers carry wide, overlapping bootstrap CIs, and paired CIs for the fusion improvement touch zero, so all orderings are indicative rather than definitive. The convex weights are dev-tuned, making dev \texttt{wconvex} numbers mildly optimistic -- though the parameter-free \texttt{mean\_prob[RoBERTa+CTD]} (test 0.650) corroborates the CTD gain without tuning. Because Ellie is wizard-controlled, ask-side timing may partly reflect prompt selection rather than participant-intrinsic behavior; response-side-only or prompt-controlled CTD variants are important future checks. Our semantic branch uses participant text only, so interviewer lexical content is excluded, but timing shortcuts remain possible and should be stress-tested. All numbers come from a single fixed split, not an independent replication, and we provide no cross-corpus, cross-lingual, or subgroup validation. The neural encoders are frozen and not fine-tuned for clinical language, which may understate their unimodal ceilings, and the acoustic detector's weakness here may be specific to this probe/cohort rather than to acoustics in general.

\section{Conclusion and Future Work}
\vspace{-1mm}
We investigated whether conversational temporal dynamics (CTD) improve dyadic depression detection on DAIC-WOZ. Reporting primarily on the development subset, a compact 24-d CTD module is the strongest single modality, and convex-weighted late fusion with semantic (and acoustic) detectors improves the best single modality from 0.746 to 0.804 dev macro-F1 and 0.631 to 0.669 test macro-F1. The gain is driven specifically by CTD: it composes with semantics, while acoustics receives zero weight in the learned three-way fusion. These results motivate treating conversational timing as a first-class, lightweight, interpretable modality in multimodal depression screening. Future work should test prompt-controlled and response-side CTD variants, joint rather than late fusion, fine-tuned or clinical-domain encoders, and external validation on additional clinical-interview corpora, including non-English and cross-cultural datasets. Before any real-world use, CTD-based screening systems should also be audited for privacy, consent, demographic fairness, and robustness to recording conditions.

\bibliographystyle{IEEEtran}
\bibliography{references}

\end{document}